\pdfoutput=1
\documentclass[11pt]{article}
\usepackage{naacl2021}

\usepackage{times}
\usepackage{latexsym}

\usepackage[most]{tcolorbox}

\usepackage{amsmath}
\usepackage{url}
\usepackage{amssymb}
\usepackage{amsfonts}
\usepackage{graphicx}
\usepackage{tabularx}
\usepackage{multirow}
\usepackage{arydshln}
\usepackage{mathtools,nccmath}

\usepackage[T5]{fontenc}

\usepackage{enumitem}

\usepackage{todonotes}

\usepackage{microtype}

\definecolor{PAT}{rgb}{0.36, 0.54, 0.66}
\definecolor{PER}{rgb}{0.82, 0.1, 0.26}
\definecolor{AGE}{rgb}{0.94, 0.87, 0.8}
\definecolor{GEN}{rgb}{1.0, 0.75, 0.0}
\definecolor{OCC}{rgb}{0.98, 0.92, 0.84}
\definecolor{LOC}{rgb}{0.55, 0.71, 0.0}
\definecolor{ORG}{rgb}{0.5, 1.0, 0.83}
\definecolor{SYM}{rgb}{0.7, 0.75, 0.71}
\definecolor{TRA}{rgb}{0.93, 0.53, 0.18}
\definecolor{DAT}{rgb}{0.57, 0.63, 0.81}

\newcommand{\annotatePAT}[1]{
\begin{tcolorbox}[colback=PAT!5!white,colframe=PAT!75!black,hbox,title=PAT,on line,before upper={\rule[-3pt]{0pt}{10pt}},boxrule=1pt,
boxsep=0pt,left=2pt,right=2pt,top=2pt,bottom=.5pt]
  #1
\end{tcolorbox}
}

\newcommand{\annotateOCC}[1]{
\begin{tcolorbox}[colback=OCC!5!white,colframe=OCC!75!black,hbox,title=OCC,on line,before upper={\rule[-3pt]{0pt}{10pt}},boxrule=1pt,
boxsep=0pt,left=2pt,right=2pt,top=2pt,bottom=.5pt]
  #1
\end{tcolorbox}
}

\newcommand{\annotateLOC}[1]{
\begin{tcolorbox}[colback=LOC!5!white,colframe=blue!75!black,hbox,title=LOC,on line,before upper={\rule[-3pt]{0pt}{10pt}},boxrule=1pt,
boxsep=0pt,left=2pt,right=2pt,top=2pt,bottom=.5pt]
  #1
\end{tcolorbox}
}

\newcommand{\annotateORG}[1]{
\begin{tcolorbox}[colback=ORG!5!white,colframe=ORG!75!black,hbox,title=ORG,on line,before upper={\rule[-3pt]{0pt}{10pt}},boxrule=1pt,
boxsep=0pt,left=2pt,right=2pt,top=2pt,bottom=.5pt]
  #1
\end{tcolorbox}
}

\newcommand{\annotateSYM}[1]{
\begin{tcolorbox}[colback=SYM!5!white,colframe=SYM!75!black,hbox,title=SYM,on line,before upper={\rule[-3pt]{0pt}{10pt}},boxrule=1pt,
boxsep=0pt,left=2pt,right=2pt,top=2pt,bottom=.5pt]
  #1
\end{tcolorbox}
}

\newcommand{\annotateDAT}[1]{
\begin{tcolorbox}[colback=DAT!5!white,colframe=DAT!75!black,hbox,title=DAT,on line,before upper={\rule[-3pt]{0pt}{10pt}},boxrule=1pt,
boxsep=0pt,left=2pt,right=2pt,top=2pt,bottom=.5pt]
  #1
\end{tcolorbox}
}
\usepackage{cuted}

\title{COVID-19 Named Entity Recognition for Vietnamese}

\author{Thinh Hung Truong, Mai Hoang Dao \and Dat Quoc Nguyen \\
        VinAI Research, Hanoi, Vietnam\\
        \normalsize{\texttt{\{v.thinhth88, v.maidh3, v.datnq9\}@vinai.io}}
        }

\begin{document}
\maketitle

\begin{abstract}
The current COVID-19 pandemic has lead to the creation of many corpora that facilitate NLP research and downstream applications to help fight the pandemic. However, most of these corpora are exclusively for English. As the pandemic is a global problem, it is worth creating COVID-19 related datasets for languages other than English. In this paper, we present the first manually-annotated COVID-19 domain-specific dataset for Vietnamese. Particularly, our dataset is annotated for the named entity recognition (NER) task with newly-defined entity types that can be used in other future epidemics. Our dataset also contains the largest number of entities compared to existing Vietnamese NER datasets. We empirically conduct experiments using strong baselines on our dataset, and find that: automatic Vietnamese word segmentation helps improve the NER results and the highest performances are obtained by fine-tuning pre-trained language models where  the monolingual model PhoBERT for Vietnamese \cite{nguyen2020phobert} produces higher results than the multilingual model XLM-R \cite{conneau2019unsupervised}. We publicly release our dataset at:  \url{https://github.com/VinAIResearch/PhoNER_COVID19}. 
\end{abstract}

\section{Introduction}
\label{intro}

As of early November 2020, the total number of COVID-19 cases worldwide has surpassed 50M.\footnote{https://www.worldometers.info/coronavirus/worldwide-graphs/\#total-cases} The world is once again hit by a new wave of COVID-19 infection with record-breaking numbers of new cases reported everyday. Along with the outbreak of the pandemic, information about the COVID-19 is aggregated rapidly through different types of texts in different languages \cite{aizawa2020system}.
Particularly, in Vietnam, text reports containing official information from the government about COVID-19 cases are presented in great detail, including de-identified personal information, travel history, as well as information of people who come into contact with the cases. The reports are frequently kept up to date at reputable online news sources, playing a significant role to help the country combat the pandemic. 
It is thus essential to build systems to retrieve and condense information from those official sources so that related people and organizations can promptly grasp the key information for epidemic prevention tasks, and the systems should also be able to adapt and sync quickly with epidemics that take place in the future.
One of the first steps to develop such systems is to recognize relevant named entities mentioned in the texts,  which is also known as the NER task.

Compared to other languages, data resources for the Vietnamese NER task are limited, including only two public datasets from the VLSP 2016 and 2018 NER shared tasks \cite{huyen2016vlsp,nguyen2018vlsp}. Here, the VLSP-2018 NER dataset is an  extension of the VLSP-2016 NER dataset with more data. These two datasets only focus on recognizing generic entities of person names, organizations, and locations in online news articles. Thus, making them difficult to adapt to the context of extracting key entity information related to COVID-19 patients. 
This leads to our work's main goals that are: (i) To develop a NER task in the COVID-19 specified domain, that potentially impacts research and downstream applications, and (ii) To provide the research community with a new dataset for recognizing COVID-19 related named entities in Vietnamese.

\begin{table*}[!t]
\centering
\resizebox{15.5cm}{!}{
    \begin{tabular}{ l | m{0.75\textwidth} }
    \hline
    \textbf{Label} & \textbf{Definition} \\
    \hline
    PATIENT\_ID & Unique identifier of a COVID-19 patient in Vietnam. An PATIENT\_ID annotation over ``X''  refers to as the X\textsuperscript{th} patient having COVID-19 in Vietnam.\\ 
    \hline
    PERSON\_NAME & Name of a patient or person who comes into contact with a patient. \\
    \hline
    AGE & Age of a patient or person who comes into contact with a patient. \\
    \hline
    GENDER & Gender of a patient or person who comes into contact with a patient.\\
    \hline
    OCCUPATION & Job of a patient or person who comes into contact with a patient.\\
    \hline
    LOCATION & Locations/places that a patient was presented at.  \\
    \hline
    ORGANIZATION & Organizations related to a patient, e.g. company, government organization, and the like, with structures and their own functions. \\
    \hline
    SYMPTOM\&DISEASE & Symptoms that a patient experiences, and diseases that a patient had prior to COVID-19 or complications that usually appear in death reports.  \\
    \hline
    TRANSPORTATION & Means of transportation that a patient used. Here, we only tag the specific identifier of vehicles, e.g. flight numbers and bus/car plates.  \\ 
    \hline
    DATE & Any date that appears in the sentence.  \\
    \hline
    \end{tabular}
}
    \caption{Definitions of entity types in our annotation guidelines. We do not annotate nested entities.}
    \label{tab:ent_descr}
\end{table*}

In this paper, we present a named entity annotated dataset with newly-defined  entity types that can be applied to future epidemics. The dataset contains informative sentences related to COVID-19, extracted from articles crawled from  reputable Vietnamese online news sites. Here, we do not consider other types of popular social media in Vietnam such as Facebook as they contain much noisy information and are not as reliable as official news sources. We then empirically evaluate strong baseline models on our dataset. Our contributions are summarized as follows:

\begin{itemize}
\setlength\itemsep{-1pt}

    \item  We introduce the first manually annotated Vietnamese dataset in the  COVID-19 domain. Our dataset is annotated with 10 different named entity types related to COVID-19 patients in Vietnam. Compared to the VLSP-2016 and VLSP-2018 Vietnamese NER datasets, our dataset has the largest number of entities, consisting of 35K entities over 10K sentences. 
    
    \item We empirically investigate strong baselines  on our dataset, including {BiLSTM-CNN-CRF}  \cite{ma2016end} and the pre-trained  language models {XLM-R} \cite{conneau2019unsupervised} and {PhoBERT} \cite{nguyen2020phobert}. We find that: (i) Automatic Vietnamese word segmentation helps improve the NER results, and (ii) The highest results are obtained by fine-tuning the pre-trained language models, where PhoBERT does better than XLM-R.
    
    \item We publicly release our dataset for research or educational purposes. 
    We hope that our dataset can serve as a starting point for future COVID-19 related Vietnamese NLP research and applications.
\end{itemize}

\section{Related work}

Most COVID-19 related datasets are constructed from two types of sources. The first one is scientific publications, including the datasets CORD-19  \cite{wang2020cord} and LitCovid \cite{chen2020keep}, that help facilitate many types of research works, such as building search engines to retrieve relevant information from scholarly articles \cite{esteva2020co,zhang2020covidex,verspoor2020covid}, question answering and  summarization   \cite{lee2020answering,dan2020caire}. Recently, \newcite{colic2020annotating} fine-tune a BERT-based NER model on the CRAFT corpus \cite{VerspoorCLWJRCFMEXBBPH12} to recognize and then normalize biomedical ontology and terminology entities in LitCovid.

The second type is social media data, particularly Tweets. COVID-19 related Tweet datasets are built for many analytic tasks such as identification of informative Tweets \cite{nguyen2020wnut}, and disinformation  detection and fact-checking \cite{FakeCovid,alam2020fighting,alsudias-rayson-2020-covid}. 
The most relevant work to ours is proposed by \citet{zong2020extracting}, that aims to extract COVID-19 events reporting test results, death cases, cures and prevention from English Tweets. As Twitter is rarely used by Vietnamese people, we could not use it for data collection. 

\section{Our dataset}
\label{dataset}

\subsection{Entity types}

We define 10 entity types with the aim of extracting key information related to COVID-19 patients, which are especially useful in downstream applications. In general, these entity types can be used in the context of not only the COVID-19 pandemic but also in other future epidemics. The description of each entity type is briefly described in Table~\ref{tab:ent_descr}. See the Appendix for entity examples as well as some notices over the entity types.

\subsection{COVID-19 related data collection}
\label{section_datacollection}

We first crawl articles tagged with "COVID-19" or "COVID" keywords from the reputable Vietnamese online news sites, including VnExpress,\footnote{\url{https://vnexpress.net}} ZingNews,\footnote{\url{https://zingnews.vn}} BaoMoi\footnote{\url{https://baomoi.com}} and ThanhNien.\footnote{\url{https://thanhnien.vn}} These articles are dated between February 2020 
and August 2020.  
We then segment the crawled news articles' primary text content into sentences using RDRSegmenter \cite{NguyenNVDJ2018} from VnCoreNLP \cite{vu-etal-2018-vncorenlp}.  

To retrieve informative sentences about COVID-19 patients, we employ BM25Plus \cite{trotman2014improvements} with search queries of common keywords appearing in sentences that report confirmed, suspected, recovered, or death cases as well as the travel history or location of the cases. From the top 15K sentences ranked by BM25Plus, we manually filter out sentences that do not contain information related to patients in Vietnam, thus resulting in a dataset of 10027 raw sentences.

\subsection{Annotation process}

We develop an initial version of our annotation guidelines and then randomly sample a pilot set of 1K sentences from the dataset of 10027 raw sentences for the first phase of annotation. Two of the guideline developers are employed to annotate the pilot set independently. Following \newcite{brandsen2020creating}, we utilize F\textsubscript{1} score to measure the inter-annotator agreement between the two annotators at the entity span level, resulting in an F1 score of 0.88.  We then host a discussion session to resolve annotation conflicts, identify complex cases, and refine the guidelines. 

In the second annotation phase, we divide the whole dataset of 10027 sentences into 10 non-overlapping and equal subsets. Each subset contains 100 sentences from the pilot set from the first annotation phase. For this second phase, we employ 10 annotators who are undergraduate students with strong linguistic abilities (here, each annotator annotates a subset, paid 0.05 USD per sentence). Annotation quality of each annotator is measured by F\textsubscript{1} calculated over the 100 sentences that already have gold annotations  from the pilot set. All annotators are asked to revise their annotations until they achieve an  F\textsubscript{1} of at least 0.92. Finally, we revisit each annotated sentence to make further corrections if needed, resulting in a final gold dataset of 10027 annotated sentences.

Note that when written in Vietnamese texts, in addition to marking word boundaries, white space is also used to separate syllables that constitute words. Therefore, the annotation process is performed at syllable-level text for convenience. 
To obtain a word-level variant of the dataset, we apply the  RDRSegmenter to perform automatic Vietnamese word segmentation, e.g. a 4-syllable written text ``bệnh viện Đà Nẵng'' (Da Nang hospital) is word-segmented into a 2-word text ``bệnh\_viện\textsubscript{hospital} Đà\_Nẵng\textsubscript{Da\_Nang}''. Here, automatic Vietnamese word segmentation outputs do not affect gold boundaries of entity mentions.

\begin{table}[!t]
    \centering
    \setlength{\tabcolsep}{0.3em}
    \resizebox{7.5cm}{!}{
    \begin{tabular}{l|l|l|l| l}
    \hline
        \textbf{Entity Type} & \textbf{Train} & \textbf{Valid.} & \textbf{Test} & \textbf{All}  \\ \hline
        PATIENT\_ID & 3240 & 1276 & 2005 & 6521 \\\hdashline
        PERSON\_NAME & 349 & 188 & 318 & 855 \\\hdashline
        AGE & 682 & 361 & 582 & 1625 \\\hdashline
        GENDER & 542 & 277 & 462 & 1281 \\\hdashline
        OCCUPATION & 205 & 132 & 173 & 510 \\\hdashline
        LOCATION & 5398 & 2737 & 4441 & 12576  \\\hdashline
        ORGANIZATION & 1137 & 551 & 771 & 2459 \\\hdashline
        SYMPTOM\&DISEASE & 1439 & 766 & 1136 & 3341 \\\hdashline
        TRANSPORTATION & 226 & 87 & 193 & 506 \\\hdashline 
        DATE & 2549 & 1103 & 1654 & 5306 \\\hline
        \# Entities in total  & 15767 & 7478 & 11735 & 34984 \\ \hline
        \# Sentences in total  &  5027 & 2000 & 3000  & 10027 \\\hline
    \end{tabular}
    }
    \caption{Statistics of our dataset.}
    \label{tab:statistic}
\end{table}

\subsection{Data partitions}

We randomly split the gold annotated dataset of 10027 sentences into training/validation/test sets with a ratio of 5/2/3, ensuring comparable distributions of entity types across these three sets. Statistics of our dataset is presented in Table~\ref{tab:statistic}.

\setcounter{table}{3}
\begin{table*}[!t]
    \centering
    \setlength{\tabcolsep}{0.3em}
    \resizebox{15.5cm}{!}{
    \begin{tabular}{ll || l | l | l | l | l | l | l | l | l | l || l | l  }
    \hline
        & \textbf{Model} & PAT. & PER. & AGE & GEN. & OCC. & LOC. & ORG. &  SYM. & TRA. & DAT. &  Mic-F\textsubscript{1} &  Mac-F\textsubscript{1}\\
        \hline 
       \multirow{3}{*}{\rotatebox[origin=c]{90}{\textbf{Syllable}}} & BiL-CRF & 0.953 & 0.855 & 0.943 & 0.947 & 0.588 & 0.915 & 0.808 & 0.801 & 0.794 & 0.976 & 0.906 & 0.858 \\
       & XLM-R\textsubscript{base} & 0.978 & 0.902 & 0.957 & 0.842 & 0.560 & 0.941 & 0.842  & 0.858 & 0.924 & 0.982 & 0.925 & 0.879  \\
        & XLM-R\textsubscript{large} & \textbf{0.982} & 0.933 & 0.962 & 0.958 & 0.692 & \textbf{0.943}  & 0.853 & 0.854 & 0.943 & 0.987 & 0.938 & 0.911  \\
        \hline
        \multirow{3}{*}{\rotatebox[origin=c]{90}{\textbf{Word}}} & BiL-CRF & 0.953 & 0.874 & 0.950 & 0.947 & 0.605 & 0.911 & 0.831 & 0.799 & 0.902 & 0.976 & 0.910 & 0.875 \\
        & PhoBERT\textsubscript{base} & 0.981 & 0.903 & 0.962 & 0.954 & 0.749 & \textbf{0.943} & 0.870 & 0.883 & 0.966 & 0.987 & 0.942 & 0.920 \\
        & PhoBERT\textsubscript{large}  & 0.980 & \textbf{0.944} & \textbf{0.967} & \textbf{0.968} & \textbf{0.791} & 0.940 & \textbf{0.876} & \textbf{0.885} & \textbf{0.967} & \textbf{0.989} & \textbf{0.945} & \textbf{0.931}   \\ 
        \hline
    \end{tabular}
    }
    \caption{Strict F\textsubscript{1} score for each entity type (denoted by its first 3 characters), and Micro- and Macro-average F\textsubscript{1} scores (denoted by Mic-F\textsubscript{1} and Mac-F\textsubscript{1}, respectively). BiL-CRF abbreviates the baseline BiLSTM-CNN-CRF. \textbf{Syllable} and \textbf{Word}  denote results obtained when using syllable- and word-level based dataset settings, respectively. }
    \label{tab:results}
\end{table*}

\section{Experiments}
\subsection{Experimental setup }
\label{benchmark}

We formulate the COVID-19 NER task for Vietnamese as a sequence labeling problem with the BIO tagging scheme.  
We conduct experiments on our dataset using strong baselines to investigate: (i) the influence of automatic Vietnamese word segmentation  (here, input sentence can be represented in either syllable or word level),  and (ii) the usefulness of pre-trained language models. 
The baselines 
include:  {BiLSTM-CNN-CRF}  \cite{ma2016end} and the pre-trained  language models {XLM-R} \cite{conneau2019unsupervised} and {PhoBERT} \cite{nguyen2020phobert}. XLM-R  is a multi-lingual variant of RoBERTa \citep{RoBERTa}, pre-trained on a 2.5TB multilingual dataset that contains 137GB of syllable-level Vietnamese texts. PhoBERT is a monolingual variant of RoBERTa, pre-trained on a 20GB word-level Vietnamese dataset. 

\setcounter{table}{2}
\begin{table}[!t]
    \centering
    \begin{tabular}{l|l}
    \hline
        \textbf{Hyper-parameter} & \textbf{Value}  \\
        \hline
        Optimizer & Adam \\
        Learning rate & 0.001 \\
        Mini-batch size & 36 \\
        LSTM hidden state size & 200 \\
        Number of BiLSTM layers & 2 \\
        Dropout & [0.25, 0.25] \\
        \hline
        \hline
        Character embedding size & 50 \\
        Filter length, i.e. window size & 3 \\
        Number of filters & 30 \\
        \hline
    \end{tabular}
    \caption{Hyper-parameters for BiLSTM-CNN-CRF.}
    \label{tab:blstm_config}
\end{table}


We employ the BiLSTM-CNN-CRF implementation from AllenNLP \cite{Gardner2017AllenNLP}. Training BiLSTM-CNN-CRF requires input pre-trained syllable- and word-level embeddings for the syllable- and word-level settings, respectively. Thus we employ the pre-trained Word2Vec syllable and word embeddings for Vietnamese from \newcite{tuan-nguyen-etal-2020-pilot}. These embeddings are fixed during training. Optimal hyper-parameters that we grid-searched for BiLSTM-CNN-CRF are presented in Table  \ref{tab:blstm_config}. 
We utilize the \textit{transformers} library \cite{wolf-etal-2020-transformers} to fine-tune XLM-R and PhoBERT for the syllable- and word-level settings, respectively, using Adam \cite{KingmaB14} with a fixed learning rate of 5.e-5 and a batch size of 32 \cite{RoBERTa}. 

The baselines are trained/fine-tuned for 30 epochs. We evaluate the Micro-average F\textsubscript{1} score after each epoch on the validation set (here, we apply early stopping if we find no performance improvement after 5 continuous epochs). We then choose the best model checkpoint to report the final score on the test set. Note that each  F\textsubscript{1} score reported is an average over 5 runs with different random seeds. 

\subsection{Main results} 

Table~\ref{tab:results} shows the final entity-level NER results of the baselines on the test set. In addition to the standard Micro-average F\textsubscript{1} score, we also report the Macro-average F\textsubscript{1} score. 

We categorize the results under two comparable settings of using syllable-level dataset and its automatically-segmented word-level variant for training and evaluation. We find that the performances of word-level models are higher than their syllable-level counterparts, showing  that automatic Vietnamese word segmentation helps improve NER, e.g. BiLSTM-CNN-CRF improves from 0.906 to 0.910 Micro-F\textsubscript{1} and from 0.858 to 0.875 Macro-F\textsubscript{1}.  

We also find that fine-tuning the pre-trained language models XLM-R and PhoBERT helps produce better performances than BiLSTM-CNN-CRF. Here, PhoBERT outperforms XLM-R (Micro-F\textsubscript{1}: 0.945 vs. 0.938; Macro-F\textsubscript{1}: 0.931 vs. 0.911), thus reconfirming the effectiveness of pre-trained monolingual  language models on the language-specific downstream tasks \cite{nguyen2020phobert}. 

\subsection{Error analysis}
\label{analysis}

We perform an error analysis using the best performing model  PhoBERT\textsubscript{large} that produces 353 incorrect predictions in total on the validation set. 

The first error group consists of \underline{69/353}  instances with correct entity boundaries (i.e. exact spans) and incorrect entity labels. It is largely due to the fact that the model could not differentiate between LOCATION and ORGANIZATION entities. 
This is not surprising because of the ambiguity between these two entity types, in which the same entity mention may act as either LOCATION or ORGANIZATION depending on the sentence context. Also, in terms of contact tracing, it would be more useful to label an organization-like entity mention as LOCATION if we can infer that a patient presented at that organization; however, such inference requires additional world knowledge about the entity. 
In addition, in this error group, the model also struggles to recognize OCCUPATION entities correctly. Recall that OCCUPATION entity mention must represent the job of a particular person labeled with PERSON\_NAME or PATIENT\_ID.
Therefore, it may cause confusion to the model for deciding whether an occupation is linked to a determined person or not in a single sentence context.

The second error group contains \underline{65/353} instances with inexact spans overlapped with gold spans but having correct entity labels. 
These errors generally happen with multi-word ORGANIZATION entity mentions, where (i) an ORGANIZATION entity contains a nested location inside its span, e.g.  ``\textit{Bệnh viện Lao và Bệnh phổi Cần Thơ}'' ({Can Tho hospital for Tuberculosis and Lung disease}; here, ``Can Tho'' is a province in Vietnam), or (ii) an organization is a subdivision of a larger organization, e.g. ``\textit{Khoa tim mạch - Bệnh viện Bạch Mai}'' ({Department of Cardiology - Bach Mai Hospital}).\footnote{Word segmentation is not shown for simplification.}  

The third group of \underline{8/353} errors with overlapped inexact spans and incorrect entity labels does not provide us with any useful insight. 
The final group of remaining \underline{211/353} errors is accounted for predicted entities corresponding with gold O labels. Particularly in the case of LOCATION, where generic mentions, such as  ``\textit{Bệnh viện tỉnh}'' (province hospital), ``\textit{Trạm y tế xã}'' (commune medical station), ``\textit{chung cư}'' (apartment), are recognized as entities, while in fact, they are not.

\section{Conclusion}
\label{conclusion}
In this paper, we have presented the first manually-annotated Vietnamese dataset in the COVID-19 domain, focusing on the named entity recognition task. We empirically conduct experiments on our dataset to compare strong baselines and find that the input representations and the pre-trained language models all have influences on this COVID-19 related NER task. We hope that our dataset can serve as the starting point for further Vietnamese NLP research and  applications in fighting the COVID-19 and other future epidemics. 

%

\bibliography{custom}
\bibliographystyle{acl_natbib}

\newpage

\section*{Appendix}

\subsubsection*{Annotation examples}

\bigskip

\begin{strip}
\rule[0.75ex]{\linewidth}{0.25pt}
\underline{Example 1:}

    Bệnh nhân "\annotatePAT{669}" là \annotateOCC{bác sĩ} làm việc tại \annotateLOC{Bệnh viện Đa khoa Đồng Nai} 
    
    Patient "\annotatePAT{669}" is a \annotateOCC{doctor} working at \annotateLOC{Dong Nai General Hospital}

\rule[0.75ex]{\linewidth}{0.25pt}
\underline{Example 2:}

     \annotateORG{Bệnh viện Bệnh Nhiệt đới TP HCM} xét nghiệm dương tính lần một đêm \annotateDAT{12/3}.
    
    \annotateORG{Ho Chi Minh City Hospital for Tropical Diseases} returns a positive test result in the evening of \annotateDAT{12/3}.

\rule[0.75ex]{\linewidth}{0.25pt}
\underline{Example 3:}

    Hai nữ điều dưỡng \annotateLOC{Bệnh viện Bạch Mai} lây từ bên ngoài và lây nhiễm cho nhau.

    Two nurses of \annotateLOC{Bach Mai Hospital} got infected from external source and then infected each other.
    
\rule[0.75ex]{\linewidth}{0.25pt}
\underline{Example 4:}

    Bệnh nhân tử vong tại \annotateLOC{Bệnh viện Phổi Đà Nẵng} với chẩn đoán \annotateSYM{viêm phổi nặng}, \annotateSYM{suy đa tạng không hồi phục}, trên bệnh nhân \annotateSYM{suy thận mạn giai đoạn cuối}.
    
    The patient died at \annotateLOC{Da Nang Lung Hospital}, diagnosed with \annotateSYM{severe pneumonia} with history of \annotateSYM{unrecoverable multiorgan dysfunction syndrome},  \annotateSYM{terminal chronic kidney failure}.
    
\rule[0.75ex]{\linewidth}{0.25pt}


Here, PAT, OCC, LOC, DAT and SYM abbreviate PATIENT\_ID, OCCUPATION, LOCATION, DATE and SYMPTOM\&DISEASE, respectively. Recall that an  annotation PATIENT\_ID over ``X''  refers to as the X\textsuperscript{th} patient having COVID-19 in Vietnam (e.g.  in Example 1: "669" refers to as the 669\textsuperscript{th} patient).

\end{strip}

\subsubsection*{Notices over entity types}

We have two principles for selecting the ten entity types: (i) Entities should contain key information related to the COVID-19 patients (here, the information should be helpful in the context of contact tracing and monitoring the growth of the pandemic); and (ii) The availability of entity types in the text, i.e., how frequent does each of the entity types appear. This is decided based on manual observations of news articles.

In the context of contact tracing, it is more useful to broaden the scope of location.
For example, when a patient is presented at an organization, we refer to that organization as a location if we can infer its specific location on the map.
In Example 1, we would label the entity mention ``\textit{Bệnh viện Đa khoa Đồng Nai}'' (Dong Nai General Hospital) with LOCATION as its provide information about the place that a patient used to be at. On the other hand, in Example 2, the entity mention ``\textit{Bệnh viện Bệnh Nhiệt đới TP HCM}'' (Ho Chi Minh City Hospital for Tropical Diseases)  is labeled as ORGANIZATION because it acts as the subject executing a specific action (i.e. reporting a  test result). 

For OCCUPATION, AGE and GENDER entities, we only tag them if we can link the corresponding entity mentions to a specific entity with NAME or PATIENT\_ID label within the same sentence.
In Example 1, ``\textit{bác sĩ}'' (doctor)  is the occupation of patient ``\textit{669}'', thus we label this mention as an entity of type OCCUPATION.
However, in Example 3, we do not label `` \textit{điều dưỡng}'' (nurses) as OCCUPATION as we cannot link this mention to any determined person.

For SYMPTOM\&DISEASE entities, we prefer the entities to be as detailed as possible. For instance, in Example 4 we consider words denoting the levels of severity as part of diseases, such as ``\textit{nặng}'' (severe), ``\textit{không hồi phục}'' (unrecoverable), ``\textit{giai đoạn cuối}'' (terminal) and ``\textit{mạn} (chronic).

\end{document}